\documentclass[10pt,twocolumn,letterpaper]{article}

\pdfoutput=1

\usepackage{amsthm}
\usepackage{colortbl}
\usepackage[pagenumbers]{cvpr}
\usepackage[acronym]{glossaries}

\makeatletter
\DeclareRobustCommand\onedot{\futurelet\@let@token\@onedot}
\def\@onedot{\ifx\@let@token.\else.\null\fi\xspace}
\def\eg{\emph{e.g}\onedot} 
\def\ie{\emph{i.e}\onedot}

\def\wrt{w.r.t\onedot} 

\makeatother

\def\svhn{SVHN\xspace}
\def\cars{Cars\xspace}
\def\faces{FACES\xspace}
\def\cifar{CIFAR-10\xspace}
\def\shapes{3D Shapes\xspace}
\def\detaux{\textbf{Detaux}\xspace}
\def\maintask{principal task\xspace}

\theoremstyle{definition}
\newtheorem{definition}{Definition}[section]
\newtheorem{prop}{Proposition}[section]

\definecolor{red}    {HTML}{b7211f}
\definecolor{orange} {HTML}{FFA500}
\definecolor{blue}   {HTML}{4169E3}
\definecolor{green}  {HTML}{147546}
\definecolor{purple} {HTML}{92268F}
\definecolor{gray}   {HTML}{606060}

\newacronym{mtl}{MTL}{Multi-Task Learning}
\newacronym{stl}{STL}{Single-Task Learning}
\newacronym{mlp}{MLP}{Multi-Layer Perceptron}
\newacronym{vae}{VAE}{Variational Auto-Encoders}

\definecolor{cvprblue}{rgb}{0.21,0.49,0.74}
\usepackage[pagebackref,breaklinks,colorlinks,allcolors=cvprblue]{hyperref}

\title{Disentangled Latent Spaces Facilitate Data-Driven Auxiliary Learning}

\author{
Geri Skenderi\textsuperscript{1}\quad
Luigi Capogrosso\textsuperscript{2}\quad
Andrea Toaiari\textsuperscript{2}\quad
Matteo Denitto\textsuperscript{3}\\
Franco Fummi\textsuperscript{2}\quad
Simone Melzi\textsuperscript{4}\\
\textsuperscript{1}Bocconi University, Bocconi Institute for Data Science and Analytics\quad
\textsuperscript{2}University of Verona\\
\textsuperscript{3}HUMATICS - SYS-DAT Group\quad
\textsuperscript{4}University of Milano-Bicocca
}

\begin{document}

\maketitle

\begin{abstract}
Auxiliary tasks facilitate learning in situations where data is scarce or the \maintask{} of interest is extremely complex.
This idea is primarily inspired by the improved generalization capability induced by solving multiple tasks simultaneously, which leads to a more robust shared representation.
Nevertheless, finding optimal auxiliary tasks is a crucial problem that often requires hand-crafted solutions or expensive meta-learning approaches.
In this paper, we propose a novel framework, dubbed \detaux{}, whereby a weakly supervised disentanglement procedure is used to discover a new unrelated auxiliary classification task, which allows us to go from a \gls{stl} to a \gls{mtl} problem.
The disentanglement procedure works at the representation level, isolating the variation related to the \maintask{} into an isolated subspace and additionally producing an arbitrary number of orthogonal subspaces, each of which encourages high separability among projections.
We generate the auxiliary classification task through a clustering procedure on the most disentangled subspace, obtaining a discrete set of labels.
Subsequently, the original data, the labels associated with the \maintask{}, and the newly discovered ones can be fed into any \gls{mtl} framework.
Experimental validation on both synthetic and real data, along with various ablation studies, demonstrates promising results, revealing the potential in what has been, so far, an unexplored connection between learning disentangled representations and \gls{mtl}.
The source code is available at \url{https://github.com/intelligolabs/Detaux}.
\end{abstract}

\glsresetall

\section{Introduction} \label{cha:cha_intro}

Human learning is often considered a combination of processes (\eg{}, high-level acquired skills, and evolutionary encoded physical perception) that are used together and can be transferred from one problem to another.
Inspired by this, \textit{\gls{mtl}}~\cite{caruana1997multitask} represents the machine learning paradigm where multiple tasks are learned together to improve the generalization ability of a model by using shared knowledge that derives from considering different aspects of the input.
Specifically, this is achieved by jointly optimizing the model's parameters across different tasks, allowing the model to learn task-specific and shared representations simultaneously.
As a result, \gls{mtl} can lead to better generalization, improved efficiency at inference time, and improved performance on individual tasks by exploiting their underlying relationships.

\begin{figure}[t!]
    \centering
    \includegraphics[width=\linewidth]{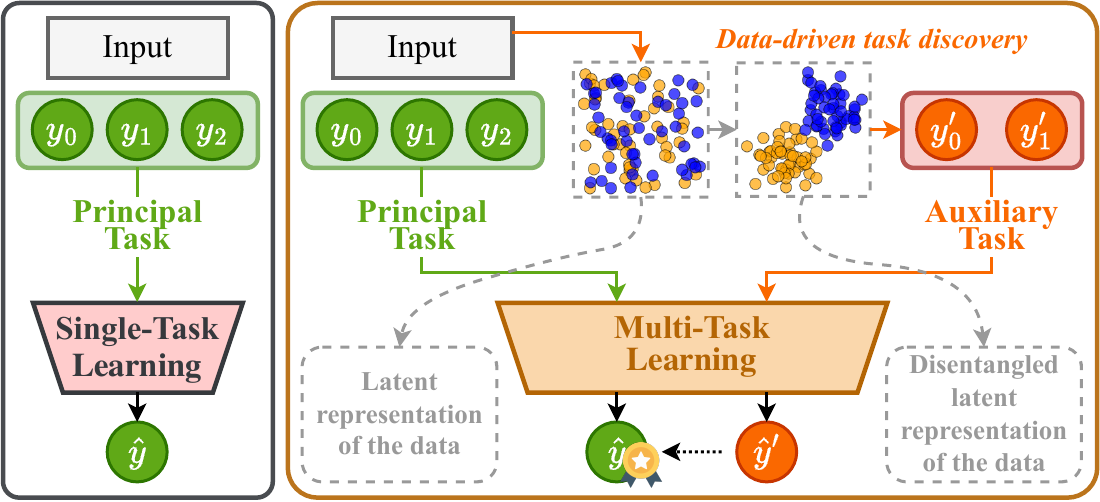}
    \caption{Overview of our data-driven auxiliary task discovery framework.
    The figure shows the difference between \gls{stl} (left) and \gls{mtl} (right) with \detaux{}, which uses an auxiliary task generated through disentanglement of the data representations to improve performance on the \maintask{}.}
    \label{fig:fig_teaser}
\end{figure}

\begin{figure*}[t!]
    \centering
    \includegraphics[width=\linewidth]{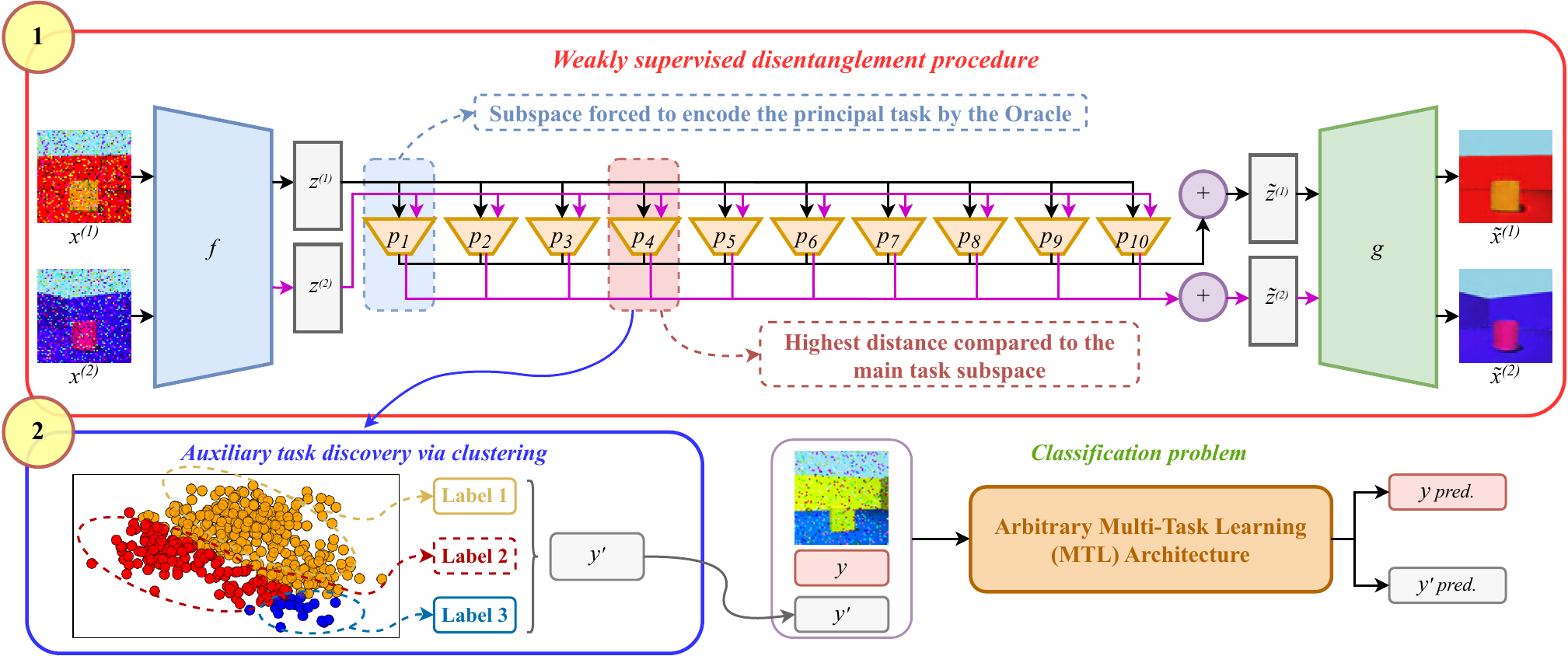}
    \caption{\detaux{} involves two steps: 
    \raisebox{.5pt}{\textcircled{\raisebox{-.9pt} {1}}} We use weakly supervised disentanglement to isolate the structural features specific to the \maintask{} in one subspace (red rectangle).
    \raisebox{.5pt}{\textcircled{\raisebox{-.9pt} {2}}} We identify the subspace with the most disentangled factor of variation related to the \maintask{}, and through a clustering module, we obtain new labels (blue rectangle).
    These can be used to create a new classification task that can be combined with the \maintask{} in \emph{any} \gls{mtl} model (bottom right part of the image).}
    \label{fig:fig_detaux}
\end{figure*}

A particular form of this learning approach, referred to as \textit{auxiliary learning}, has garnered considerable interest in recent years~\cite{liebel2018auxiliary}. 
Auxiliary learning consists of using an additional set of tasks, dubbed auxiliary tasks. 
These tasks operate on the same input data and lead to a shared representation useful to boost the performance on the \maintask{}, \ie{}, the only task of interest.
At the state-of-the-art, auxiliary tasks are generated by meta-learning~\cite{liu2019self,navon2021auxiliary}, but this requires an a priori definition of the hierarchy of the desired auxiliary tasks and is often computationally inefficient. Thus, the question is: can we discover without prior knowledge one or more additional auxiliary tasks from the data to improve the performance on the \maintask{}?

In this paper, we explore this problem by proposing \detaux{}, a weakly supervised strategy that discovers auxiliary classification tasks that enable solving a single-task classification problem in a multi-task fashion, as depicted in Figure~\ref{fig:fig_teaser}. Specifically, \detaux{} is capable of individuating unrelated auxiliary tasks: unrelatedness in \gls{mtl} means to have two or more tasks whose features have \emph{no} semantic intersection, and proved to be effective in the literature~\cite{wang2003facial,zhou2011hierarchical,paredes2012exploiting,jayaraman2014decorrelating,zheng2019exploiting,liu2019encoding}.

Our method takes roots in the idea of~\cite{paredes2012exploiting}, where two groups of tasks, the \maintask{} and the auxiliary tasks, are given and known to be unrelated, and assumes the claim that joint learning of unrelated tasks can improve the performance on the \maintask{}. It proposes to generate a shared low-dimensional representation for both the \maintask{} and the unrelated auxiliary tasks, forcing these two representations to be orthogonal.

The procedure from~\cite{paredes2012exploiting} exploits a linear classifier and requires the labels for both the \maintask{} and the auxiliary tasks to be known.
Our method aims to follow a similar process while not requiring full supervision and using non-linear parametrizations through neural networks.
Specifically, it generates auxiliary tasks so that their labels implicitly drive an \gls{mtl} network to understand the unrelatedness between the tasks.
Our idea is to work in a specific representation space, a product manifold, to reveal the auxiliary tasks for a given \maintask{}.
We get inspiration from~\cite{fumero2021learning}, who discovered the product manifold as a convenient representation basis for disentanglement.
In particular, as shown in Figure~\ref{fig:fig_detaux}, we first extract task-specific features using a weakly supervised disentanglement procedure that implements projections on orthogonal subspaces of the latent representation; then, we identify a subspace where the respective projections are maximally separated. 
Finally, we generate new labels via a clustering module to enable integration with the primary task in any \gls{mtl} model.

In particular, this makes the proposed pipeline agnostic to the choice of the \gls{mtl} model, given that the latter acts directly on the primary and generated auxiliary labels, as shown in the bottom right of Figure~\ref{fig:fig_detaux}.
In this way, any \gls{mtl} model can be chosen depending on several factors besides performance, such as efficiency, scalability, and resource constraints. In the experimental section, we test three different \gls{mtl} models with the \detaux{} framework on four different image datasets, revealing its flexibility.

\section{Related Work} \label{cha:cha_related}

\subsection{MTL and Auxiliary Learning}

\gls{mtl}, \ie{}, the procedure through which multiple learning problems are solved at the same time~\cite{caruana1997multitask}, proved effective in reducing inference time, reaching improved accuracy, and increasing data efficiency~\cite{standley2020tasks}.
When the adopted dataset contains annotations for multiple tasks, the challenges to face concern which tasks may work well together~\cite{zamir2018taskonomy,standley2020tasks,fifty2021efficiently} or how to weigh the losses of different tasks~\cite{kendall2018multi} to create a better joint optimization objective.
Numerous methods have recently emerged that address the simultaneous resolution of multiple tasks~\cite{caruana1997multitask,gao2019nddr,vandenhende2020mti}.

A different problem arises when we would like to use an \gls{mtl} method, but the given dataset contains annotations for only one task.
Auxiliary learning aims to maximize the prediction performance on a \maintask{} by supervising the model to learn other tasks, as shown in~\cite{liu2019self,navon2021auxiliary}.
Therefore, auxiliary tasks are tasks of minor interest, or even irrelevant compared to the \maintask{} we want to solve, and thus can be seen as regularizers~\cite{liebel2018auxiliary}.
For example,~\cite{paredes2012exploiting} suggested that the use of two unrelated groups of tasks, where one of them hosts the \maintask{}, can lead to better performance, where unrelated means that an orthogonal set of features defines the two groups of tasks.
In~\cite{liebel2018auxiliary}, the authors used seemingly unrelated tasks to help the learning on one \maintask{}, this time without imposing any constraint on the feature structure.
\detaux{} exploits a product manifold space, which has already been shown by~\cite{fumero2021learning} to be effective in separating embedding subspaces that are orthogonal by design.

Moreover, some techniques leveraged meta-learning to select the most appropriate auxiliary tasks or even autonomously create novel ones.
Both~\cite{liu2019self} and~\cite{li2021meta} train two neural networks simultaneously: a label-generation model to predict the auxiliary labels and a multi-task model to train the primary task alongside the auxiliary task.
In contrast with our approach, these require the a priori definition of a hierarchy binding the auxiliary labels to the \maintask{} labels and present conflicting ideas on the possible semantic interpretation of the generated labels.
Furthermore, they are often computationally inefficient: meta-learning is a resource-intensive technique that requires retraining the entire architecture to change the employed multi-task method.
\cite{nam2023stunt} also used meta-learning, presenting a novel framework to generate new auxiliary objectives and address the niche problem of few-shot semi-supervised tabular learning.
Finally, \cite{dery2022aang} proposed to deconstruct existing natural language processing objectives within a unified taxonomy, identifying connections between them, and generating new ones by selecting the best combinations from a Cartesian product of the available options.

\subsection{Learning Disentangled Representations}

Representing data in a space where different components are independent is a long-standing research topic in machine learning.
The rise of deep learning, which relies on learning representations, has made this concept even more relevant and useful in understanding the latent space~\cite{bengio2013representation}.

Recent literature has proposed several characterizations of disentanglement, whether that is in terms of group theory~\cite{higgins2018towards}, metric and product spaces~\cite{fumero2021learning}, or permutations of element-wise, nonlinear functions~\cite{horan2021unsupervised}.
~\cite{higgins2017beta} demonstrated that variational auto-encoders can learn to disentangle by enforcing the ELBO objective, while~\cite{chen2016infogan} relied on generative adversarial networks and an information-theoretic view of disentanglement.
Later works, such as~\cite{eastwood2018framework,singh2019finegan,ojha2020elastic}, extensively explored different directions and use cases.
~\cite{locatello2019challenging} showed that completely unsupervised disentanglement is not possible due to the inability of the models to identify factors of variation.
Soon after, the authors proposed weak supervision and access to a few labels to bypass this limitation~\cite{locatello2020weakly,locatello2020labels}.
In \detaux{}, we place ourselves in the same setting of~\cite{fumero2021learning} but control and force the disentanglement by supervision only on the known \maintask{}.

\subsection{MTL and Disentanglement}

~\cite{meng2019representation} reported a connection between disentangled representations and \gls{mtl}, showing that disentangled features can improve the performance of \gls{mtl} networks, especially on data with previously unseen properties. 
Disentanglement is obtained by adversarial learning, forcing the encoded features to be minimally informative about irrelevant tasks. 
In this case, the tasks to be disentangled are known a priori, while in our case, only the \maintask{} task is known. 

\cite{yang2022factorizing} proposed a novel concept called ``Knowledge Factorization''. 
Exploiting the knowledge contained in a pre-trained \gls{mtl} network (called teacher), the idea is to train disentangled \gls{stl} networks (called students) to reduce the computational effort required by the final single-task network. 
The factorization of the teacher's knowledge is dual: they provide structural factorization and representation factorization. 
In structural factorization, they split the net into a common knowledge network and a task-specific network based on mutual information. 

Finally,~\cite{maziarka2023relationship} explores the degree of disentanglement of \gls{mtl} models in a controlled, semi-synthetic setting.
Initially, a set of task labels is created using a randomly initialized \gls{mlp} starting from the latent factors of parametric disentanglement datasets~\cite{3dshapes18}.
The authors successively train a separate neural network to solve these artificially created tasks and understand how disentangled the representations are \wrt{} the original latent factors.

In this work, we show that disentanglement in a representation space can be used as a general prior for \gls{mtl}.
After using disentanglement to mine for auxiliary tasks, an \gls{mtl} model extracts a model-specific embedding that takes advantage of the combination of the principal and newly discovered labels, improving downstream performance on the \maintask{}.

\section{Mathematical Background} \label{cha:cha_backgorund}

\subsection{Disentanglement Framework} \label{sec:sec_dis_theory}
At a high level, disentangled representation learning aims to learn a representation of the data where different latent factors are represented independently of the others; that is, we have a factorized representation~\cite{bengio2013representation}.

There are different ways to properly formalize this general concept.
In this work, we rely on the definition and approach of disentanglement proposed by~\cite{fumero2021learning}.
The primary assumption behind this framework is the manifold hypothesis, \ie{}, that high-dimensional data lies near a lower-dimensional manifold.
Building upon this idea and assuming that independent factors generate the data, it becomes reasonable to see the manifold as a product manifold: $\mathcal{M}=\mathcal{M}_{1}\times{}\mathcal{M}_{2}\times{}\ldots{}\times{}\mathcal{M}_{k}$.
In such a topological structure, each $\mathcal{M}_{i},i\in{}\{1\ldots{}k\}$, is orthogonal to the others, and thus, we would like it to represent at most one latent factor of the data.
This concept is adequately formalized by relying on the topological construct of a metric space and employing what we call a weak isometry between the data and the learned product manifolds, defined as follows.

\begin{definition}[Product Manifold Disentanglement \cite{fumero2021learning}] \label{def:def_disentanglement}
Let $\mathcal{M} =\mathcal{M}_1\times{}\mathcal{M}_2\times{}\ldots{}\times{}\mathcal{M}_k$ be the data product manifold, embedded in a high-dimensional space $\mathcal{X}$.
Furthermore, let us assume that we have access to some metric that endows these two spaces with the properties of a metric space.
A representation $z$ in some product space $\mathcal{Z}=S_1\times{}\ldots{}\times{}S_k$, such that $dim(\mathcal{Z})\ll{}dim(\mathcal{X})$, is \emph{disentangled} with respect to $\mathcal{M}$ if there exists a diffeomorphism (a bijection with a smooth inverse) $\tilde{g}:\mathcal{Z}\to{} \mathcal{M}$ such that $\forall{} x_1,x_2\in{}\mathcal{M};\forall{}i\in{}{1,\ldots{},k}$:
 \begin{align*}
    & d_{\mathcal{M}_i}(x_1^i,x_2^i)>0\implies{}d_{\mathcal{S}_i}(s_1^i,s_2^i)>0\;,\\
    & d_{\mathcal{M}_i}(x_1^i,x_2^i)=0\implies{}d_{\mathcal{S}_i}(s_1^i,s_2^i)=0\implies{}s_1^i=s_2^i\;,
\end{align*}
where $x_j^i$ is the projection of $x_j$ on $\mathcal{M}_i$ and $s_j^i=\Pi_i\tilde{g}^{-1}(x_j)$ with $\Pi_i$ being the projection onto the subspace $\mathcal{S}_i\subset{}\mathcal{Z}$.
\end{definition}

According to Definition~\ref{def:def_disentanglement}, given a pair of data $(x_{1},x_{2})$ known to differ in the $h$-th latent factor only, their learned representations are considered fully disentangled if they have fixed projections in all submanifolds $\{\mathcal{M}_{i}\}_{i=1}^{k}$, except for the $h$-th.

To provide a pictorial understanding of the above definition, consider the simple case where the data live in $M=\mathbb{R}^2=\mathbb{R}\times{}\mathbb{R}$, embedded in an ambient space $\mathcal{X}$ of arbitrary (but finite) dimension.
We can see this data manifold as the Cartesian plane and label the two submanifolds as the well-known $x$ and $y$ axes.
Given that both are diffeomorphic to open subsets of the real number line, our goal is to learn a latent representation where the $x$-coordinate is embedded into one subspace and the $y$ coordinate into the other, such that they remain separate.
In this simple example, the disentangled representation would correspond to an intuitive change in the basis in $\mathbb{R}^2$.
Very similarly, any product manifold composed of $n$ 1D, connected, non-compact submanifolds without boundary, could be represented in latent space as $\mathbb{R}^n$ while respecting Definition~\ref{def:def_disentanglement}.
This approach comes with a great advantage when it comes to its application in generating auxiliary tasks, which is that each submanifold can have a different dimensionality, as long as the sum of the dimensions of the submanifolds equals that of the product manifold.
Therefore, we can generalize the intuitive idea of an ``axis of variation'' and look for auxiliary tasks in a higher-dimensional space instead of being limited to 1D representation axes as in \gls{vae}-based methods~\cite{higgins2017beta,kim2018disentangling}.

\subsection{Disentanglement Training Procedure} \label{sec:sec_dis_training}

In practice, we consider a finite-dimensional normed vector space $\mathbf{Z}\subseteq{}\mathbb{R}^{d}$, containing the disentangled latent representation ($\mathbf{Z}$ is a particular case of a smooth manifold).
The disentangled latent representation takes the form of a Cartesian product space $\mathbf{Z}=\mathcal{S}_{1}\times{}\mathcal{S}_{2}\times{}\ldots{}\times{}\mathcal{S}_{k}$, such that $\forall{}i,j\in{}\{1{}\ldots{}k\}$, with $i\neq{}j$, $\mathcal{S}_{i}\cap{}\mathcal{S}_{j}=\{0\}$.
As mentioned above, each subspace generalizes the notion of an ``axis of variation''.
An encoder network $f: \mathcal{M} \to{} \mathbf{Z}$ produces the latent product space, and a set of non-linear operators $\{p_i\}^k_{i=1}$ projects the latent representation to the respective (latent) subspace.
The representations in each subspace are then aggregated, and a decoder $g$ maps the resulting vectors back to the input data space.
More specifically, each $\mathbf{S}_i\subseteq{}\mathcal{R}^{d}$ is defined in such a way that it has the same ambient dimensionality as the product space $\mathbf{Z}$.
Using a specific regularization (defined in Equation~\ref{eq:eq_sparsity_loss}), each subspace will have only a few non-zero entries, and the non-zero entries in one subspace will be zero in the others.
This encourages orthogonal and sparse representations for each $\mathbf{S}_i$, which can then be summed to produce a latent code.
Subsequently, this latent code is fed into a decoder $g$ that approximates the inverse of $f$.
Thus, the decoder is the approximation of the function $\tilde{g}$ in Definition~\ref{def:def_disentanglement}.

To wrap up, the representation framework operates in the following way: 
\begin{equation}
    x \xrightarrow{f} z \xrightarrow[i=1\ldots{}k]{\{p_i\}} \{s_i\} \xrightarrow[i=1\ldots{}k]{\sum_i} \tilde{z} \xrightarrow{g} \tilde{x}\;,
\end{equation}
where $\tilde{z}$ and $\tilde{x}$ are the aggregated latent representation and the reconstructed input, respectively.
The visual representation of this process is shown in Figure~\ref{fig:fig_detaux}, within the red rectangle.
In the following, we describe how it is possible to parameterize this framework with neural networks and train it end-to-end.

An autoencoder architecture is used to approximate $f$ and $g$.
The encoder $f$ receives non-i.i.d. data pairs $(x^{(1)},x^{(2)})$ and produces the latent representations $(z^{(1)},z^{(2)})$, while the decoder $g$ approximates the inverse of $f$.
The reason for training with input pairs is to have a sampling procedure designed to induce weak supervision, requiring a pair of images known to vary in at least one latent factor (this is crucial to later isolate the change from $x^{(1)}$ to $x^{(2)}$ in one subspace).
Furthermore, a set of $k$ neural networks $p_{i},i\in{}\{1\ldots{}k\}$ called projectors are trained simultaneously to map the latent codes in the subspaces $\{\mathcal{S}_{i}\}_{i=1}^{k}$, each of which contains the corresponding submanifold $\{\mathcal{M}_{i}\}_{i=1}^{k}$.

An initial warm-up phase trains $f$ and $g$ only to minimize the data reconstruction error, which is needed to learn the global data manifold $\mathcal{M}$.
After this warm-up phase, four differentiable constraints that regard different aspects of the desiderata defined in Section~\ref{sec:sec_dis_theory} are added, posing the following optimization problem:
\begin{equation} \label{eq:eq_disentanglement_loss}
    \mathcal{L}=\mathcal{L}_{rec}+\beta_{1}(\mathcal{L}_{dist}+\mathcal{L}_{spar})+\beta_{2}\mathcal{L}_{cons}+\beta_{3}\mathcal{L}_{reg}\;,
\end{equation}
where $\beta_{1}$, $\beta_{2}$, and $\beta_{3}$ are Lagrange multipliers.
In the following paragraphs, we provide only a synthesized textual description of the losses due to the lack of space and the fact that the formalizations of these constraints are not directly relevant to the description of our method.
For completeness, all formulas and additional details can be found in the Appendix (Section~\ref{supsec:supsec_dis_losses}).

$\mathcal{L}_{rec}$ corresponds to a common \textit{reconstruction loss}, implemented in practice as the squared error between the input and reconstructed images.

The \textit{distance loss}, $\mathcal{L}_{dist}$, is a contrastive loss term.
It is built based on an oracle function $\mathcal{O}:\mathbf{Z}\times{}\mathbf{Z}\to{}\{1,\ldots{}, k\}$, which calculates the subspace $\mathcal{S}_i$ where the projections of the images in the pair $(x_1,x_2)$ differ the most, in terms of their distance in the latent space.
Then, it encourages the projection representation of the two input images onto the subspaces not selected by $\mathcal{O}$ to be as close as possible while pushing the representations in $\mathcal{S}_i$ to be further apart.
In combination with $\mathcal{L}_{cons}$, it encourages the weak isometry in Definition~\ref{def:def_disentanglement}.
The oracle function is a crucial part of our method, and the version implemented in \detaux{} will be defined in Equation~\ref{eq:eq_forced_oracle}.

$\mathcal{L}_{spar}$ is an $L1$ constraint which promotes sparsity and orthogonality between the subspaces, by encouraging each one to have a few non-zero entries that will be zero in the others.
In our finite-dimensional setting, this constraint is equivalent to imposing that the product space is a direct sum of the subspaces, thus allowing the summation operation to aggregate the subspaces.

$\mathcal{L}_{cons}$, namely the \textit{consistency loss}, encourages each projector $p_{i}$ to be invariant to changes in subspaces $\mathcal{S}_{j},j\neq{}i$.
Together with $\mathcal{L}_{dist}$, this constraint encourages the metric definition of disentanglement in Definition~\ref{sec:sec_dis_theory}.

Finally, the \textit{regularization loss} $\mathcal{L}_{reg}$ introduces a penalty that ensures that the choice of the oracle $\mathcal{O}$ is uniformly distributed among subspaces to avoid the collapse of information.
This is necessary given the initial warm-up period with only the reconstruction loss being active, as there is no guarantee that information will be equally spread out among the subspaces.

\section{Methodology} \label{cha:cha_method}

\paragraph*{Setting and notation.}
We assume the existence of a labeled image dataset $D=\{\,(x^{(i)},y^{(i)})\,|\,\forall{i}\in{}\{1\dots{}N\},\,x^{(i)}\in{}\mathcal{R}^{w\times{}h\times{}c},\, y^{(i)}\in{}\mathbb{N}\}$, where $w$ is the width, $h$ the height, $c$ the number of channels, and $N$ the number of tuples (image, label).
We consider the classification task whose fundamental objective is to learn a mapping from the image space $\{x^{(i)}|\forall{i}\in{}\{1\dots{}N\}\}$ to the corresponding label $\{y^{(i)}|\forall{i}\in{}\{1\dots{}N\}\}$.

\subsection{The Principal Task-Based Oracle} \label{sec:sec_task_based_oracle}

When applied to our setting, a major drawback of the procedure proposed by~\cite{fumero2021learning} is that the oracle will assign the representations of data points with a different \maintask{} label to an arbitrary subspace at random.
To automatically discover auxiliary tasks, we must have a way to accommodate the known variation of the \maintask{} in an arbitrary subspace and fix it there.
To achieve this, we define a \maintask{}-based oracle $\hat{\mathcal{O}}:\mathbf{Z}\times{}\mathbf{Z}\to{}\{1,\ldots{}, k\}$, which ensures that the $\alpha{}$-th subspace will contain all the variation in the data corresponding to pairs $(x^{(1)},x^{(2)})$ whose elements differ in their \maintask{} label.
Note that we do not inject direct knowledge of these labels, but only whether or not they differ.
To do this, we select a subspace $\alpha{}\in{}\{1\ldots{}k\}$ where we wish to force the variation of the \maintask{} labels and define $\hat{\mathcal{O}}$ as:
\begin{equation} \label{eq:eq_forced_oracle}
    {\hat{\mathcal{O}}}(z^{(1)},z^{(2)})=
    \begin{cases}
    \alpha & \text{if } y^{(1)}\ne{}y^{(2)} \\
    \underset{i \in \{1, \ldots, k\} \setminus \alpha}{\operatorname{argmax}} d(s_{i}^{(1)}, s_{i}^{(2)}) & \text{otherwise}
    \end{cases}
\end{equation}
where $d(s_{i}^{(1)}, s_{i}^{(2)})$ is the distance between the projections of $(z^{(1)},z^{(2)})$ in the $i$-th subspace $\mathcal{S}_{i}$.
Our new oracle implies that the distance and regularization losses will always force the variation in the data to be encoded in $\mathcal{S}_{\alpha{}}$ \text{if }$y^{(1)}\ne{}y^{(2)}$, and otherwise in a different subspace.
The choice of the subspace for the case $y^{(1)}=y^{(2)}$ is made by looking at where the distance between the projections is maximal, as this is where the difference between the pair in that latent factor will be encoded.
Thanks to the consistency loss, the remaining subspaces can encode other variations while remaining invariant to the ones related to the \maintask{} and contained in $\mathcal{S}_{\alpha{}}$.
The combination of these constraints will lead us to discover a proper representation in which unknown tasks correspond to (possibly) multiple orthogonal subspaces to those of the \maintask{}.
We set $\alpha{}=1$ in practice, as this is a simple and intuitive choice, but any other value $\in \{1...k\}$ is perfectly suitable.

In practice, $\hat{\mathcal{O}}$ needs to be differentiable in order to train the disentanglement network end-to-end.
To this end, we approximate the $\operatorname*{arg\,max}$ operator by applying a $\operatorname*{soft\,max}$ at low temperature to the Euclidean distance matrix normalized by the average length of the vector representation in each subspace. This produces a discrete probability distribution over the distances, which can then be used to weigh the contributions of the projections in each subspace.

\subsection{Auxiliary Task Discovery} \label{sec:sec_aux_task_discovery}
In the disentangled representation of the input data, where the known \maintask{} variation is encoded into a subspace of our choice, we look for new auxiliary tasks in the remaining subspaces.
Intuitively, we wish to have a disentangled subspace that exhibits a clustering tendency over the projected data.
This notion is implicitly built into the disentanglement loss function in Equation~\ref{eq:eq_disentanglement_loss}, mainly due to the distance loss $\mathcal{L}_{dist}$.
Let $\mathcal{S}_{j}$ (obtained from the projector $p_{j}$) be the subspace where $\mathcal{L}_{dist}$ is minimal after training.
We apply a clustering algorithm to the latent representation lying in $\mathcal{S}_{j}$, as shown inside the blue rectangle in Figure~\ref{fig:fig_detaux}.
After clustering, we obtain a set of discrete pseudo-labels, determining the new auxiliary classification task.
Given that the disentanglement procedure already indicates how much each subspace might contain different clusters (via $\mathcal{L}_{dist}$), choosing the subspace $\mathcal{S}_{j}$ makes it so that the image embeddings are already well separated, providing a big advantage for the clustering procedure.

Although it is possible to use an arbitrary clustering algorithm, we would like it to support clusters of arbitrary shapes and not to directly specify the number of clusters.
Therefore, we utilize HDBSCAN~\cite{campello2013density}, as it allows us to cluster data points based on their proximity and density without explicitly specifying the number of clusters.
This algorithm can associate points that cannot be assigned to any cluster to a ``noise'' cluster, which we can retain as an additional label of the auxiliary task, making it very robust. 
If HDBSCAN finds only one cluster, we denote the run as unsuccessful and stop the procedure,  as training it would lead to trivial MTL results.
Otherwise, we have discovered a novel task and its corresponding labels $y'\in{}\mathbb{N}$, which can be used with any \gls{mtl} model, as shown in the bottom right part of Figure~\ref{fig:fig_detaux}.
In this work, we limit ourselves to finding only one auxiliary task.
Scaling on more tasks is the subject of future work.
At this stage, we have enriched our dataset with an additional set of labels, obtaining $D'=\{\,(x_{i},y_{i},y'_{i})\,|\,\forall{}i\in{}\{ 1\dots{}N\},\,x_{i}\in{}\mathcal{R}^{w\times{}h\times{}c},\,y_{i},y'_{i}\in{}\mathbb{N}\}$.

\paragraph*{Theoretical analysis.}
Under the assumption that tasks living in orthogonal spaces help increase \gls{mtl} performance~\cite{paredes2012exploiting}, we now show why our method regularizes the learning procedure and implicitly guides it towards orthogonal feature spaces for each task.
For the rest of the paragraphs, we assume perfect disentanglement, \ie{}, $\mathcal{L}=0$ in Equation~\ref{eq:eq_disentanglement_loss}.
Let $\mathbf{X}\in{}\mathbb{R}^{N\times{}d}$ be the vectorized representation of the dataset ($d=w\times{}h\times{}c)$, $\mathcal{S}_{\alpha}\in{}\mathbb{R}^{N\times{}n}$ be the subspace that contains the representation of the \maintask{}, forced by Equation~\ref{eq:eq_forced_oracle}, and $\mathcal{S}_{j}\in{}\mathbb{R}^{N\times{}m}$ be the subspace that contains the representation of the auxiliary task, as described previously.
We then obtain the following results:

\begin{prop} \label{prop:prop_uncorrelated}
The representations $\mathcal{S}_{\alpha}$ and $\mathcal{S}_{j}$ are not correlated.
Furthermore, given the respective distance matrices $\mathbf{\Delta}_{ab}^{\alpha}=||s_{\alpha}^{(a)}-s_{\alpha}^{(b)}||_2$, $\mathbf{\Delta}_{ab}^{j}=||s_{j}^{(a)}-s_{j}^{(b)}||_2$, $a,b\in{}\{ 1, \dots{}, N\}$, and some scalar $\gamma{}\in{}\mathbb{R}$, we have that $\mathbf{\Delta}^{j}\neq{}\gamma{} \mathbf{\Delta}^{\alpha}$. 
\end{prop}

\begin{prop} \label{prop:prop_grams}
Let the overlap matrix of two square matrices $\mathcal{A}$ and $\mathcal{B}$ be defined as $\mathcal{V}_{AB}=\mathcal{A}^T\mathcal{B}$.
Then, the overlap matrix between the eigenvectors of the Gram matrices $\mathcal{G}_{\alpha}=\mathcal{S}_{\alpha}\mathcal{S}_{\alpha}^T$ and $\mathcal{G}_{j}=\mathcal{S}_{j}\mathcal{S}_{j}^T$ is different from the Identity matrix $\mathcal{I}_{n}$, \ie{}, they have different eigenvectors.
\end{prop}

The proofs are deferred to Section~\ref{supsec:supsec_theoretical_analysis_proof} in the Appendix.
Specifically, Proposition~\ref{prop:prop_uncorrelated} implies that the relationship between $d(s^{(a)}_{\alpha{}},s^{(b)}_{\alpha{}})$ will not influence the one between $d(s^{(a)}_{j},s^{(b)}_{j})$, given that the information encoded in each subspace is different.
Furthermore, due to Proposition~\ref{prop:prop_grams}, the structure of the pairwise similarity between points (given by the eigenvectors of the Gram matrices) is different in the two subspaces.
Thus, a clustering algorithm that relies on pairwise similarity, such as HDBSCAN, will produce different clusterings.

\section{Experiments} \label{cha:experiments}

\paragraph{Implementation details.}
Our code is written within the PyTorch Lightning framework.
We fix the batch size to 32, the learning rate to $0.0005$, and use the AdamW~\cite{loshchilov2017decoupled} optimizer.
The disentanglement model is trained for 40 epochs on \shapes{}~\cite{3dshapes18} and 400 epochs on \faces{}~\cite{ebner2010faces}, \cifar{}~\cite{krizhevsky2009learning}, \svhn{}~\cite{netzer2011reading}, and \cars{}~\cite{krause20133d}.
The first quarter of the epochs is used as a warm-up period where only $\mathcal{L}_{rec}$ is active.
The multipliers $\beta_{1}, \beta_{2}, \beta_{3}$ follow an exponential warm-up routine after the reconstruction-only phase, so that the constraints they control are gently introduced in the optimization procedure.
The projectors $p_{i},i\in{}\{1\ldots{}k\}$ are implemented as two layers \glspl{mlp}.
Finally, all the \gls{mtl} models are trained for 150 epochs. All experiments are performed on NVIDIA RTX 3090 GPUs.

\begin{table*}[t!]
    \centering
    \caption{Classification accuracy on the \faces{}~\cite{ebner2010faces}, \cifar{}~\cite{krizhevsky2009learning}, \svhn{}~\cite{netzer2011reading}, and \cars{}~\cite{krause20133d} datasets.
    (*) The results are the ones reported in the original paper since we encountered challenges in replicating the performance.
    In \textbf{bold}, the best results. 
    \underline{Underlined} the second best.
    In parentheses, the change in performance over \gls{stl}.}
    \begin{tabular}{l|c|c|c|c}
\toprule 
\textbf{Learning Paradigm} & \textbf{\faces{}~\cite{ebner2010faces} $\uparrow$} & \textbf{\cifar{}~\cite{krizhevsky2009learning} $\uparrow$} & \textbf{\svhn{}~\cite{netzer2011reading} $\uparrow$} & \textbf{\cars{}~\cite{krause20133d} $\uparrow$} \\
\midrule
\rowcolor{gray!10} STL                          & 0.915 & 0.844 & \underline{0.956} & 0.711 \\
\midrule
MAXL~\cite{liu2019self}               & 0.933 \small{\color{green}{(+0.018)}} & 0.868 \small{\color{green}{(+0.024)}} & 0.953 \small{\color{red}{(-0.003)}} & 0.638 \small{\color{red}{(-0.073)}}\\
AuxiLearn~\cite{navon2021auxiliary}   & 0.915 \small{(+0.000)} & 0.811 \small{\color{red}{(-0.033)}} & 0.943 \small{\color{red}{(-0.013)}} & 0.644* \small{\color{red}{(-0.067)}}\\
\midrule
MTL-HPS~\cite{caruana1997multitask} + \detaux{} \textbf{(ours)} & \underline{0.951 \small{\color{green}{(+0.036)}}} & 0.848 \small{\color{green}{(+0.004)}} & 0.954 \small{\color{red}{(-0.002)}} & \underline{0.789 \small{\color{green}{(+0.078)}}}\\
NDDR~\cite{gao2019nddr} + \detaux{} \textbf{(ours)}             & 0.932 \small{\color{green}{(+0.017)}} & \underline{0.872 \small{\color{green}{(+0.028)}}} & 0.952 \small{\color{red}{(-0.004)}} & 0.712  \small{\color{green}{(+0.001)}}\\
MTI~\cite{vandenhende2020mti} + \detaux{} \textbf{(ours)}       & \textbf{0.978 \small{\color{green}{(+0.063)}}} & \textbf{0.910 \small{\color{green}{(+0.066)}}} & \textbf{0.961 \small{\color{green}{(+0.005)}}} & \textbf{0.807 \small{\color{green}{(+0.096)}}}\\
\bottomrule
\end{tabular}
    \label{tab:tab_detaux_real_data_results}
\end{table*}

\subsection{Synthetic Data} \label{sec:sec_exp_synth_data}
To showcase the capabilities of \detaux{}, we begin our experimental validation with the \shapes{} dataset, a common benchmark in the disentanglement literature~\cite{kim2018disentangling,locatello2019disentangling,fumero2021learning}.
\shapes{} comprises six generative factors: floor hue, wall hue, object hue, scale, shape, and orientation. 
It is parametrically generated through the Cartesian product between these factors, resulting in 480,000 images.
To adapt it to our case, we treat the classification of one generative factor as the \maintask{} and pretend that we do not know the others.

Due to the synthetic nature of the images in \shapes{}, solving a classification task with a neural network can be trivial, leaving a limited possibility for improvement through \gls{mtl}.
Specifically, using a simple VGG16~\cite{simonyan2014very} model, we achieve perfect accuracy in each of the six possible tasks.
Thus, to render this setting slightly more complicated, we add salt-and-pepper noise to 15\% of the image pixels.
With the presence of noise, the classification of the object scale (4 classes) becomes challenging.
Hence, we have chosen it as the primary task for our experiments.
The number of subspaces $k$ is set to 10 as in~\cite{fumero2021learning}.

As described in Section~\ref{sec:sec_aux_task_discovery}, we cluster the most disentangled subspace (not considering the one dedicated to the \maintask{}) according to the disentanglement loss.
The HDBSCAN minimum cluster size hyperparameter is set to 2\% of the number of data points $N$.
In this experiment, the subspace chosen for the clustering seems to coincide with the one encoding the information regarding the object hue (10 classes).
In fact, given the optimal disentanglement on \shapes{}, the auxiliary labels generated by the clustering procedure almost perfectly match the ground-truth object hue label, having homogeneity and completeness scores of 0.999.

We then feed the noisy \shapes{} images and the enriched label set into an \gls{mtl} hard parameter-sharing architecture with a VGG16~\cite{simonyan2014very} as the backbone and compare \gls{stl} vs \gls{mtl}.
For this comparison, we need to perform a train-test split on \shapes{}, which is non-trivial since the possible combinations of the latent factors in the dataset are present exactly once.
Therefore, we split the dataset based on the floor and wall hue labels, allocating the images that contain 5 out of the 10 values for both factors only to the testing set, resulting in a 75-25 train-test split.
On the \maintask{}, \gls{mtl} achieves an accuracy of \textbf{0.889}, outperforming the 0.125 obtained by \gls{stl} by a large margin, \ie{},  \textbf{\color{green}{+0.746}}.

\subsection{Real Data} \label{sec:sec_real_data}
As in the previous example, during the disentanglement procedure, pairs of images are sampled only based on the \maintask{} labels.
In the multi-class \faces{} dataset, this corresponds to the person's facial expression.
In \cifar{}, \svhn{}, and \cars{}, it corresponds to the only annotated labels.
We used an encoder-decoder architecture based on a ResNet-18~\cite{he2016deep} to obtain a high-quality reconstruction.
The number of subspaces $k$ is set to 10.
During the auxiliary task discovery, we set the HDBSCAN minimum cluster size hyperparameter to 1\% of the number of data points $N$ for all datasets.

We compare our approach with two different auxiliary learning methods, \ie{}, MAXL~\cite{liu2019self} and AuxiLearn~\cite{navon2021auxiliary}.
Unlike these auxiliary learning architectures that exploit a meta-learning procedure, our discovered auxiliary task can be employed interchangeably with any \gls{mtl} model.

To have as much control over the experiments as possible and focus on the benefits of our discovered auxiliary task, we choose parameter-sharing \gls{mtl} networks to ensure that the gains are due to the new task and not to the specific architecture or advanced learning dynamics.
Modern \gls{mtl} research has mostly shifted towards more complex ideas, which is why these well-known approaches are dated before 2021.
This dichotomy is also discussed in~\cite{lin2023libmtl}.
Thus, we select three different models: the standard Hard Parameter Sharing for \gls{mtl} (MTL-HPS)~\cite{caruana1997multitask}, weighing the losses to give more importance to the \maintask{}, as suggested in~\cite{kendall2018multi}, NDDR~\cite{gao2019nddr}, and MTI~\cite{vandenhende2020mti}.
All these models have a loss term composed as a summation of each task's classification loss.
In this way, during the backpropagation, the gradient alters any shared parameters between the two tasks while looking to maximize performance on both, which is what drives the improved generalization capability.

In particular, MTI was proposed to operate with an HRNet backbone~\cite{wang2020deep}.
This type of network performs multi-resolution fusion, starting from a high-resolution convolution stream and gradually adding high-to-low-resolution convolution streams one by one.
Since the datasets on which we operate contain mainly images of size $\leq 224\times{}224$, learning HRNet from scratch results in low-quality representations.
Therefore, we use the official MTI code, which uses an HRNet pre-trained on ImageNet~\cite{deng2009imagenet}.

Table~\ref{tab:tab_detaux_real_data_results} summarizes the results.
MTI, with our generated auxiliary labels, displays the best performance.
Furthermore, even simpler \gls{mtl} models, such as MTL-HPS and NDDR, achieve superior results compared to MAXL and AuxiLearn.
Most notably, we outperform \gls{stl} with at least one of the \gls{mtl}+\detaux{} models in all the datasets, whereas MAXL and AuxiLearn have large performance discrepancies between the datasets.

For completeness, we report that we exploit pre-trained backbones for the \gls{mtl} models on the \cars{} dataset, which contains very complex images and is categorized as a fine-grained classification dataset.
For the disentanglement phase with this dataset, we change the encoder $f$ so that it does not produce a dense representation in the bottleneck layer but a compressed feature map.
Thus, latent space projectors are learned by using $1\times{}1$ convolution, and disentanglement losses are applied to the flattened feature map.

\subsection{Research Questions} \label{sec:sec_ablation}

\paragraph{How important is the disentangled representation for auxiliary task discovery?}
This experiment aims to show how disentanglement effectively extracts task labels from the underlying data structure.
On the \faces{} dataset, we compare the auxiliary task generated by \detaux{} with the auxiliary task resulting from the clustering on the latent space of an autoencoder that only learns to reconstruct.
Without disentanglement, MTL-HPS can only reach 0.9 accuracy, worse than the 0.915 obtained by STL.
This reveals that performing auxiliary task mining on the entangled autoencoder space provides a less informative auxiliary task to the \gls{mtl} network compared to our approach.
We provide further qualitative evidence of this observation in Figure~\ref{fig:fig_disentanglement_for_auxiliary_task_discovery} of the Appendix. 

\paragraph{Is it possible to use other clustering algorithms?}
\begin{table}[t!]
    \centering
    \caption{Classification accuracy on the \faces{}~\cite{ebner2010faces} and \cars{}~\cite{krause20133d} datasets when using different clustering algorithms to generate the auxiliary task labels of \detaux{}.
    All the \gls{mtl} results are obtained using the MTL-HPS~\cite{caruana1997multitask} model.
    In parentheses, the change in performance over \gls{stl}.}
    \resizebox{\columnwidth}{!}{
    \begin{tabular}{l|c|c|c}
\toprule 
\textbf{} & \textbf{Clustering} & \textbf{\faces{}~\cite{ebner2010faces} $\uparrow$} & \textbf{\cars{}~\cite{krause20133d} $\uparrow$} \\
\midrule
\rowcolor{gray!10} STL & --               & 0.915 & 0.711\\
\rowcolor{gray!10} MTL & HDBSCAN~\cite{campello2013density}          & 0.951 & 0.789\\
\midrule
MTL & KMeans~\cite{lloyd1982least}        & 0.953 \small{\color{green}{(+0.038)}} & 0.789 \small{\color{green}{(+0.078)}}\\
MTL & KMeans++~\cite{arthur2006k}         & 0.934 \small{\color{green}{(+0.019)}} & 0.790 \small{\color{green}{(+0.079)}}\\
MTL & MeanShift~\cite{comaniciu2002mean}  & 0.963 \small{\color{green}{(+0.048)}} & 0.783 \small{\color{green}{(+0.072)}}\\
\bottomrule
\end{tabular}}
    \label{tab:tab_detaux_clustering_results}
\end{table}

An immediate question that may come to mind regarding \detaux{} is its flexibility with regard to the clustering method.
As mentioned in Section~\ref{sec:sec_aux_task_discovery}, we rely on HDBSCAN due to its nice properties.
We aim to show that our pipeline can improve downstream performance even with other (and simpler) clustering algorithms.
In particular, we use two versions of the K-Means algorithm~\cite{lloyd1982least,arthur2006k}, which assume a flat geometry, and MeanShift~\cite{comaniciu2002mean}, which works well even in non-flat geometries.
We set $k$ equal to the number of clusters found by HDBSCAN and estimate the bandwidth parameter of MeanShift for each dataset~\cite{comaniciu2002mean}.
The results are presented in Table~\ref{tab:tab_detaux_clustering_results} and clearly show that our method is flexible to the choice of the clustering algorithm, improving performance in all cases compared to \gls{stl}.

\vspace{-0.3cm}
\paragraph{Are the generated labels correlated?}
\begin{table}[t!]
    \centering
    \caption{Normalized and Adjusted Mutual Information between the principal and auxiliary task labels generated using \detaux{}.}
    \resizebox{\columnwidth}{!}{
    \begin{tabular}{l|c|c}
\toprule 
\textbf{Dataset} & \textbf{Normalized MI $\downarrow$} & \textbf{Adjusted MI $\downarrow$} \\
\midrule
\faces{}~\cite{ebner2010faces}         & 0.1405 & 0.1390\\
\cifar{}~\cite{krizhevsky2009learning} & 0.0033 & 0.0031\\
\svhn{}~\cite{netzer2011reading}       & 0.0033 & 0.0030\\
\cars{}~\cite{krause20133d}            & 0.0311 & 0.0085\\
\bottomrule
\end{tabular}}
    \label{tab:tab_detaux_mutual_info_results}
    \vspace{-0.5cm}
\end{table}

To empirically verify that our pipeline design aligns with the underlying theoretical analysis, we calculate the Normalized (NMI) and Adjusted (AMI) Mutual Information between the \maintask{} and the auxiliary task labels generated by \detaux{}.
These metrics are used in the clustering literature to measure the agreement between two label assignments, independently of the order~\cite{romano2016adjusting}.
The results in Table~\ref{tab:tab_detaux_mutual_info_results} indicate that the two sets of labels are almost uncorrelated, with the labels from \faces{} exhibiting minimal correlation. 

To further confirm our claim, we run a contingency-based $\chi^2$ test with the null hypothesis that the two groups have no significant differences.
For all datasets, the p-values are essentially 0 (the largest one being $9.06\times{}10^{-17}$), allowing us to reject the null hypothesis with high confidence.

\section{Conclusion} \label{cha:cha_conclusions}

In this paper, we propose a novel perspective on the utility of disentangled representations, utilizing them as a proxy for auxiliary learning in order to improve the accuracy of a \maintask{}, originally solvable only in a \gls{stl} fashion. 
Our proposed pipeline facilitates the weakly supervised discovery of new tasks from a factorized representation.
These newly discovered tasks can be incorporated into any \gls{mtl} framework, and we empirically show that this offers improved performance on the \maintask{}.
\\~\\
\noindent{\textbf{Acknowledgments.}} Geri Skenderi is funded by the European Union through the Next Generation EU - MIUR PRIN PNRR 2022 Grant P20229PBZR.
Furthermore, this study was also carried out within the ``COMMIS5.0: Chatbot e interfacce umano-centriche per l'OttiMizzazione della Manifattura e l'Innovazione nei Servizi 5.0'' Funded by PR Veneto FESR 2021-2027, Obiettivo Specifico 1.1, Azione 1.1.1 ``Rafforzare la ricerca e l’innovazione (in collaborazione) tra imprese e organismi di ricerca'' Sub A ``Rafforzare la ricerca e l’innovazione tra imprese e organismi di ricerca''.

{
    \small
    \bibliographystyle{ieeenat_fullname}
    \bibliography{03_bibi}
}

\glsresetall

\clearpage \appendix \section*{Appendix}

\section{Disentanglement Loss Functions} \label{supsec:supsec_dis_losses}
In this section, we formally present the loss functions utilized to enforce the latent product manifold structure and promote disentanglement, as defined in Section~\ref{sec:sec_dis_training} of the main paper.
The loss function is the following.
\begin{equation}
    \mathcal{L}=\mathcal{L}_{rec}+\beta_{1}(\mathcal{L}_{dist}+\mathcal{L}_{spar})+\beta_{2}\mathcal{L}_{cons}+\beta_{3}\mathcal{L}_{reg}\;.
\end{equation}

$\mathcal{L}_{rec}$ corresponds to a \textit{reconstruction loss}, implemented in practice as the squared error between the input and the reconstructed images following the subspaces' aggregation operation in the latent space.
It is defined as:
\begin{gather}
    \mathcal{L}_{rec}=\|x-\bar{x}\|_2^2\;,\\
    \bar{x} = g \biggl({\sum}( p_{1}(f(x)), \ldots{}, p_{k}(f(x))\biggr)\;.
\end{gather}
This term is necessary to learn the global structure of the manifold $\mathcal{M}$.

The \textit{distance loss}, $\mathcal{L}_{dist}$, is a contrastive loss term that follows the oracle $\mathcal{O}$, defined in Section~\ref{sec:sec_dis_theory}, which calculates the subspace $\mathcal{S}_i$ where the projections of the images in the pair $(x_1, x_2)$ differ the most, and encourages the projection representation of the two input images onto the subspaces not selected by $\mathcal{O}$ to be as close as possible.
It is defined as:
\begin{equation} \label{eq:eq_distance_loss}
    \mathcal{L}_{dis}=\sum_{i=1}^{k} (1 -\lambda_i) \delta_{i}^2+ \lambda_i\max(m-\delta_{i},0)^2\;,
\end{equation}
where $\lambda_i= 1 $ if $\mathcal{O}(z^{(1)},z^{(2)})=i$ and $0$ otherwise, while $m$ is a hyperparameter that constrains the points to be at least a distance $m$ from each other.

$\mathcal{L}_{spar}$ is a \textit{L1} constraint which promotes sparsity and orthogonality between the subspaces.
It is defined as:
\begin{equation} \label{eq:eq_sparsity_loss}
    \mathcal{L}_{spar} = \sum_{i=1}^{k}\| p_{i}(f(x))\odot \sum_{j\neq i}^k p_{j}(f(x)) \|_1\;.
\end{equation}
This constraint allows the disentanglement framework to use the sum operation to aggregate the subspaces.
The minimization of $\mathcal{L}_{spar}$ promotes sparsity and orthogonality between the subspaces, encouraging each one to have a few non-zero entries that will be zero in the others.
In our finite-dimensional setting, this loss is equivalent to imposing that the product space is a direct sum of the subspaces.

$\mathcal{L}_{cons}$, namely the \textit{consistency loss}, encourages each projector $p_{i}$ to be invariant to changes in subspaces $\mathcal{S}_{j},j\neq{}i$.
It is defined as:
\begin{equation} \label{eq:eq_consistency_loss}
    \mathcal{L}_{cons}=\sum_{i=1}^{k}||p_{i}(f_\theta(\hat{x}_{s_i})) - s_i||_2^2\;,
\end{equation}
with $s_i = p_if(x_1)$ and $ \hat{x}_{s_i} = g ({\sum}( p_i f(x_1),p_{j \neq i}f(x_2)))$.
Along with $\mathcal{L}_{dist}$, this constraint encourages a metric definition of disentanglement, \ie{}, given a pair of images that are different in image space \wrt{} to a particular factor, they should be equally different in the latent representation of that attribute, hosted only in one submanifold which composes the global, product manifold of the latent representation.

Finally, the \textit{regularization loss} $\mathcal{L}_{reg}$ introduces a penalty that ensures that the choice of the oracle $\mathcal{O}$ is uniformly distributed among subspaces to avoid the collapse of information.
This is necessary given the initial warm-up period with only the reconstruction loss being active, as there is no guarantee that information will be equally spread out among the subspaces.
It is defined as:
\begin{equation}
    \mathcal{L}_{reg}= \sum_{j=1}^k \left(\frac{1}{N}\sum_{n=1}^N \mathbf{A}_{n,j} - \frac{1}{k} \right)^2\;,
\end{equation}
with $\mathbf{A} \in \mathbb{R}^{N\times k}$ being the practical implementation of the oracle indicator variables of Equation~\ref{eq:eq_distance_loss} in a batch of $N$ pairs, obtained by applying a weighted softmax to the distance matrix of pairs in each of the $k$ subspaces.

\section{Proofs} \label{supsec:supsec_theoretical_analysis_proof}
We provide here the proofs for both Proposition~\ref{prop:prop_uncorrelated} and Proposition~\ref{prop:prop_grams} that are present in Section~\ref{sec:sec_aux_task_discovery} of the main paper.
The idea behind them is to rely on the assumption of perfect disentanglement.
For the sake of clarity and to make this supplementary material self-contained, we restate the assumptions and the propositions from scratch.

Under the assumption that tasks living in orthogonal spaces help increase \gls{mtl} performance~\cite{paredes2012exploiting}, we now show why our method regularizes the learning procedure and implicitly guides it towards orthogonal feature spaces for each task.
For the rest of the paragraphs, we assume perfect disentanglement \ie{}, $\mathcal{L}=0$ in Equation~\ref{eq:eq_disentanglement_loss}.
Let $\mathbf{X}\in{}\mathbb{R}^{N\times{}d}$ be the vectorized representation of the dataset ($d=w\times{}h\times{}c)$, $\mathcal{S}_{\alpha}\in{}\mathbb{R}^{N\times{}n}$ be the subspace that contains the representation of the \maintask{}, forced by Equation~\ref{eq:eq_forced_oracle}, and $\mathcal{S}_{j}\in{}\mathbb{R}^{N\times{}m}$ be the subspace that contains the representation of the auxiliary task, as described previously. In the following, we assume that $n=m$, which is also what we implement in practice.
We then obtain the following results:

\begin{prop}
The representations $\mathcal{S}_{\alpha}$ and $\mathcal{S}_{j}$ are not correlated.
Furthermore, given the respective distance matrices $\mathbf{\Delta}_{ab}^{\alpha}=||s_{\alpha}^{(a)}-s_{\alpha}^{(b)}||_2$, $\mathbf{\Delta}_{ab}^{j}=||s_{j}^{(a)}-s_{j}^{(b)}||_2$, $a,b\in{}\{ 1, \dots{}, N\}$, and some scalar $\gamma{}\in{}\mathbb{R}$, we have that $\mathbf{\Delta}^{j}\neq{}\gamma{}\mathbf{\Delta}^{\alpha}$. 
\end{prop}

\begin{proof}
The uncorrelatedness of the representations is a direct consequence of the complete minimization of the sparsity and orthogonality constraint $\mathcal{L}_{spar}$ (Equation~\ref{eq:eq_sparsity_loss}).
For any vector $x\in{}\mathbf{X}$, we have that:
\begin{equation}
    \| p_{\alpha}(f(x))\odot p_{j}(f(x)) \|_1 = 0\;,
\end{equation}
which directly implies that $\langle \mathcal{S}_{\alpha} , \; \mathcal{S}_{j} \rangle_F = 0$.
Assuming without loss of generality that the representations are centered at $0$, this leads to the conclusion that the two representations are uncorrelated as they have $0$ covariance.

Similarly, the second part of the proposition is a direct consequence of the complete minimization $\mathcal{L}_{spar}$ and the consistency constraint $\mathcal{L}_{cons}$ (Equation~\ref{eq:eq_consistency_loss}).
The complete minimization of $\mathcal{L}_{cons}$ makes the nonlinear operator $p_i$ invariant to changes in the subspaces $S_j, \forall{} j \neq i.$~\cite{fumero2021learning}.
Now, we proceed by contradiction.
Assume that the two distance matrices are proportional to a scalar multiple of each other.
Then, having proportional pairwise distances would imply that there exists a linear function $\omega: \mathbf{Z} \longrightarrow\ \mathbf{Z}$ $p_j(f(x)) = \omega(p_{\alpha}(f(x)))$, implying that the vector in the auxiliary subspace is a function of $p_{\alpha}(f(x))$.
A straightforward example of this would be a permutation followed by scaling.
If this were the case, $p_j(\cdot)$ would not be invariant to the changes in $S_{\alpha}$, as it directly depends on it, so by contradiction, we can conclude that $\mathbf{\Delta}^{j}\neq{}\gamma{} \mathbf{\Delta}^{\alpha}$.
\end{proof}

\begin{prop}
Let the overlap matrix of two square matrices $\mathcal{A}$ and $\mathcal{B}$ be defined as $\mathcal{V}_{AB}=\mathcal{A}^T\mathcal{B}$.
Then, the overlap matrix between the eigenvectors of the Gram matrices $\mathcal{G}_{\alpha}=\mathcal{S}_{\alpha}\mathcal{S}_{\alpha}^T$ and $\mathcal{G}_{j}=\mathcal{S}_{j}\mathcal{S}_{j}^T$ is different from the Identity matrix $\mathcal{I}_{n}$, \ie{}, they have different eigenvectors.
\end{prop}

\begin{proof}
From Proposition~\ref{prop:prop_uncorrelated}, we have that $\mathcal{G}_{\alpha} \not\propto \mathcal{G}_{j}$, meaning that the pairwise similarities in both spaces are not proportional (intended as equal or different up to a scalar factor).
Given that both Gram matrices are Symmetric and Positive Semi-Definite, by the Spectral Theorem \cite{strang2022introduction}, we can diagonalize them and obtain a set of $n$ orthonormal eigenvectors with real eigenvalues:
\begin{align}
    \mathcal{G}_{\alpha} = \mathcal{U}_{\alpha} \Lambda_{\alpha} \mathcal{U}_{\alpha}^T\;,\\
    \mathcal{G}_{j} = \mathcal{U}_{j} \Lambda_{j} \mathcal{U}_{j}^T\;.
\end{align}

The reason why we are interested in these eigenvectors is that they contain orthogonal directions of pairwise similarity, thus indicating the pairwise groupings present in the dataset.
By simply calculating the overlap matrix on the above eigendecomposition, it is straightforward to see that:
\begin{align*}
    \mathcal{V}_{\mathcal{G}_{\alpha}\mathcal{G}_{j}} = (\mathcal{U}_{\alpha} \Lambda_{\alpha} \mathcal{U}_{\alpha}^T)^T \mathcal{U}_{j} \Lambda_{j} \mathcal{U}_{j}^T\;. \\
    \: = \mathcal{U}_{\alpha} \Lambda_{\alpha} \mathcal{U}_{\alpha}^T \mathcal{U}_{j} \Lambda_{j} \mathcal{U}_{j}^T\;. \\
    \; \neq \mathcal{I}_{n}\;.
\end{align*}
Therefore, the eigenvectors do not perfectly align, and thus, the fixed directions of pairwise similarities are different in the two subspaces.
\end{proof}

\section{Additional Research Questions} \label{supsec:supsec_additional_research_questions}

\paragraph{Why return to image space for \gls{mtl}?}
One may ask why we did not work directly in the latent feature space found by the disentanglement procedure.
We did some preliminary experiments in this direction, but they yielded inconclusive results and raised implementation issues that are beyond the scope of this paper. 
A reason is that most \gls{mtl} frameworks for image classification require convolution, which is not well-defined for feature vectors living in the latent space.
Another reason is that \detaux{} works at a representation level, regardless of any classification aim induced by a specific classification framework. 
Its sole purpose is to reveal, together with the subspace dedicated to the \maintask{} determined by the initial labels, other orthogonal complementary subspaces, which can be assumed as tasks if they admit clustering.
The output of \detaux{} is an enriched set of labels that can be exploited with any \gls{mtl} model. 
In addition, \detaux{} allows us to visualize and interpret the disentangled subspaces since it reconstructs the images. 
This procedure allowed us to understand that, in the toy example on \shapes{}~\cite{3dshapes18}, the additional task corresponds to the object's hue (one of the generative factors). 
Unfortunately, in more complex real cases, clear interpretation becomes more challenging, barely revealing the gender as an additional task in the \faces{}~\cite{ebner2010faces} benchmark.
In the other cases, we had no clue.
However, it should be noted that we focused on producing a framework that transforms a single task classification problem into an \gls{mtl} one.
We left eventual interpretability analyses for future work.

\paragraph{Why only use a single auxiliary task?}
In our experiments, we always use a single auxiliary task extracted from the most disentangled subspace (excluding the one allocated for the \maintask{}).
We made this choice to be able to test our research question - can disentanglement help us discover at least one subspace from which to extract a good auxiliary task? - while keeping the presentation of the various stages of the pipeline as straightforward as possible.
Furthermore, the number of additional tasks places a non-trivial computational burden on the parameter-sharing models we implement for \gls{mtl}.
The scalability of such models is an interesting research direction, which we believe is beyond the scope of this work.
Hence, the use of more auxiliary tasks is deferred to future work. 

\paragraph{Are the \gls{mtl} results statistically significant?}
To empirically validate whether the results presented in Table~\ref{tab:tab_detaux_real_data_results} are statistically significant, we focus on the \svhn{}~\cite{netzer2011reading} dataset, where only MTI + \detaux{} outperforms the \gls{stl} baseline.
Therefore, we compare with the best competitor, MAXL~\cite{liu2019self}, over five different seeds.
For comparison, we conducted a two-sample t-test on the results to check if the means are significantly different from each other.
Considering a significance level of 0.05, we obtain a p-value of 0.0004, which confirms that the results are significant.
On average (over these five runs), MTI + \detaux{} reports a 1.1\% gain in accuracy compared to MAXL.

\paragraph{What does disentanglement look like from a qualitative perspective?}
\begin{figure}[t!]
    \centering
    \includegraphics[width=\linewidth]{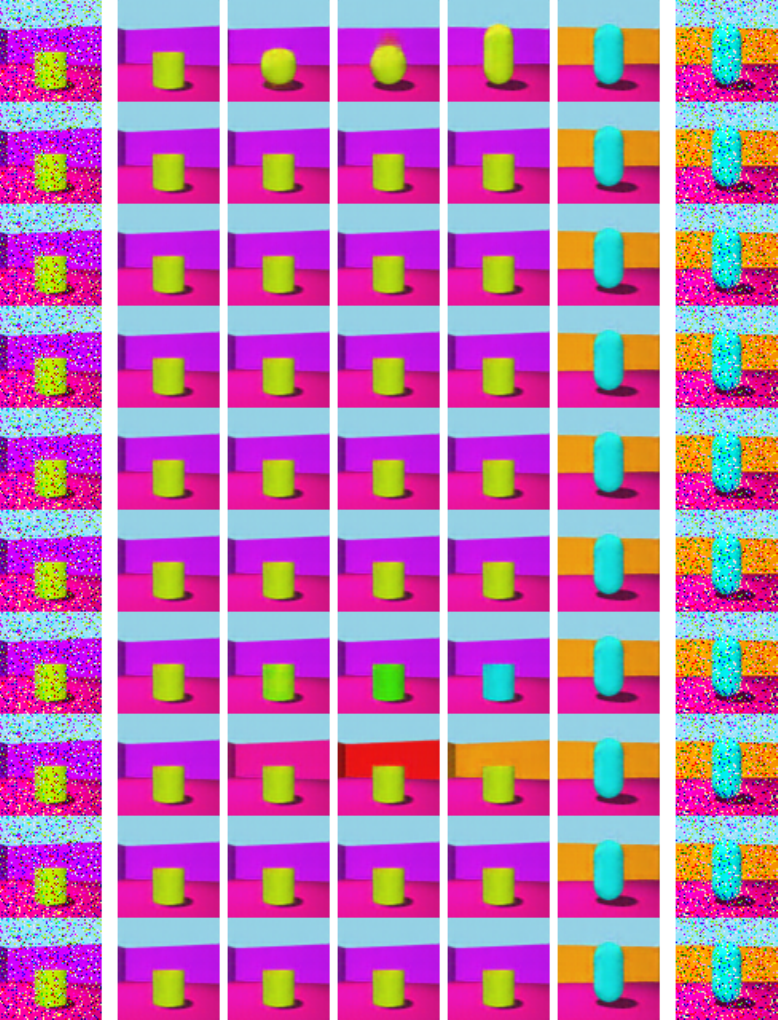}
    \caption{Visual interpretation of the disentanglement procedure on \shapes{} with noisy input.
    The disentanglement model factorizes the representation and forces the \maintask{} (\ie{}, the object scale) in the first subspace, albeit with other factors of variation.
    The remaining factors (floor, wall, and object hues) are all disentangled in different subspaces and can be used to discover additional auxiliary tasks. 
    Best viewed in color.}
    \label{fig:fig_3d_shapes_disentanglement}
\end{figure}

\begin{figure}[t!]
    \centering
    \includegraphics[width=\linewidth]{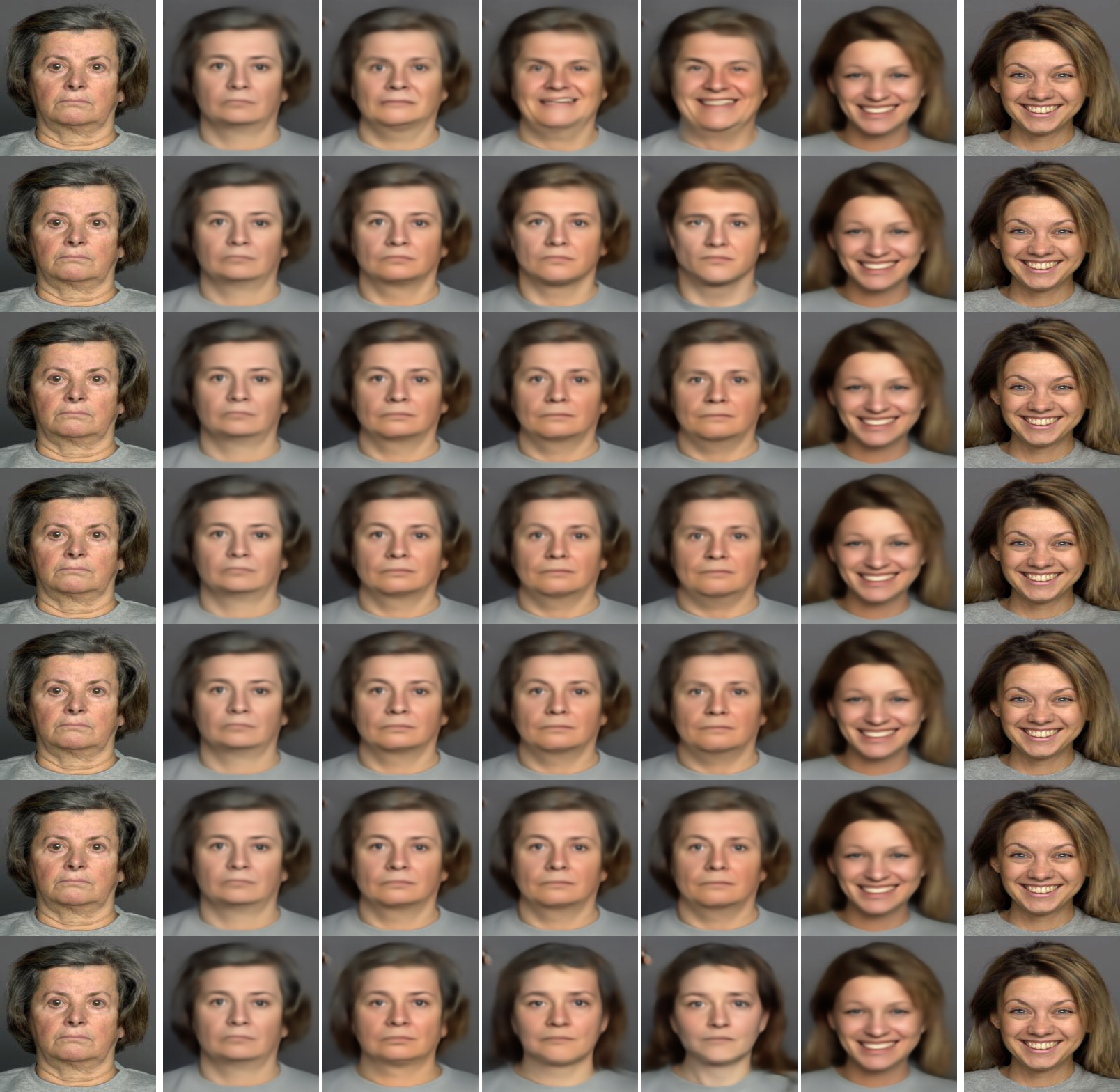}
    \caption{Visual interpretation of the disentanglement on \faces{}.
    With real data, it becomes more difficult to factorize and visualize true generative factors.
    The naked eye can definitely realize that the variations between the image pair are contained in different subspaces.
    The first row shows the effect of our supervised oracle, which forces the \maintask{} (\ie{}, the person’s facial expression) in the first subspace.
    At the same time, other variations arise in the other subspaces, allowing us to mine for auxiliary tasks.} 
    \label{fig:fig_faces_disentanglement}
\end{figure}

Figure~\ref{fig:fig_3d_shapes_disentanglement} and Figure~\ref{fig:fig_faces_disentanglement} provide a qualitative perspective of the disentanglement on the \shapes{} and \faces{} datasets.
In both visualizations, the outermost columns (far left and far right) represent the two images composing an input image pair, respectively.
The adjacent columns (near the outermost) depict the reconstruction of the images.
The three central columns display variations corresponding to specific factors encoded in individual subspaces.
This is done by linearly interpolating between the representations of the pair and then reconstructing the result.
Each row highlights a distinct subspace, showcasing how different generative factors are disentangled and isolated for targeted analysis.

In Figure~\ref{fig:fig_3d_shapes_disentanglement}, we can see how the disentanglement model factorizes the representation and forces the \maintask{} (\ie{}, the object scale) in the first subspace, although with other factors of variation.
Furthermore, we can see that setting a higher number of subspaces than generative factors is not an issue, since it is possible for the model to collapse the variation in certain subspaces.

Figure~\ref{fig:fig_faces_disentanglement} shows how the disentanglement procedure behaves when used on real data.
Specifically, it becomes clear that it is more difficult to factorize and visualize true generative factors.
One can notice how only the eyes and mouth, related to smiling and being happy, are altered, while the rest of the face remains almost identical.
In the second row, we can see a candidate auxiliary task, where the subject's gender seems to change and display different traits.
These traits are indeed diverse from the ones dealing with the change in emotion, isolated in the first subspace, showing how we can extract orthogonal auxiliary tasks.

\paragraph{Is disentanglement crucial for auxiliary task discovery? (cont.d)}
\begin{figure}[t!]
   \centering
   \includegraphics[width=\linewidth]{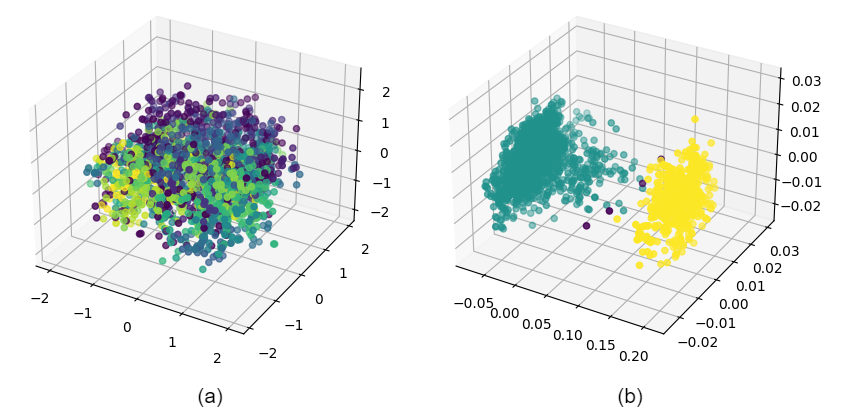}
   \caption{3D visualization of the discovered auxiliary task in the entangled autoencoder feature space \textit{(a)} and the most disentangled subspace \textit{(b)}, on the \faces{} dataset.
   The high-dimensional representations are projected to 3D space using Principal Component Analysis (PCA).
   Different colors mean different clusters found by HDBSCAN.
   The representation in \textit{(a)} is highly entangled, while the one in the disentangled representation space \textit{(b)} displays a clear and reasonable grouping.
   Best viewed in color.}
   \label{fig:fig_disentanglement_for_auxiliary_task_discovery}
\end{figure}

Figure~\ref{fig:fig_disentanglement_for_auxiliary_task_discovery} highlights from a qualitative point of view the importance of disentanglement in discovering auxiliary tasks.
The visualizations show the feature spaces of the \faces{} dataset in two different settings.
Subfigure \textit{(a)} illustrates the entangled feature space, where representations remain highly mixed, leading to less discernible clusters.
Conversely, subfigure \textit{(b)} depicts the most disentangled subspace, where the features are clearly grouped into distinct and interpretable clusters.
The high-dimensional feature representations are reduced to 3D space using PCA, and the clusters are identified using the HDBSCAN~\cite{campello2013density} algorithm.
The evident separation in the disentangled subspace underscores its importance for auxiliary task mining.
These results also emphasize from a qualitative point of view that disentanglement not only simplifies representation learning but also facilitates structured auxiliary task discovery.

\end{document}